\documentclass[journal]{IEEEtran}

\usepackage[utf8]{inputenc}
\usepackage{color}
\usepackage{xcolor}
\usepackage{array}
\usepackage{verbatim}
\usepackage{float}
\usepackage{amsmath}
\usepackage{amsthm}
\usepackage{amssymb}
\usepackage{graphicx}
\usepackage{longtable}
\usepackage{multirow}
\usepackage{booktabs}
\usepackage{balance}
\providecommand{\tabularnewline}{\\}
\usepackage{algorithmic}
\usepackage{longtable}

\floatstyle{ruled}
\newfloat{algorithm}{tbp}{loa}
\providecommand{\algorithmname}{Algorithm}
\floatname{algorithm}{\protect\algorithmname}
\usepackage{multirow}
\usepackage[unicode=true,
bookmarks=false,
breaklinks=false,pdfborder={0 0 1},colorlinks=false]
{hyperref}
\hypersetup{
	colorlinks,bookmarksopen,bookmarksnumbered,citecolor=blue,urlcolor=blue}
\usepackage{cite}

\usepackage{lipsum}
\usepackage{mathtools}
\usepackage{cuted}

\floatstyle{ruled}
\newfloat{algorithm}{tbp}{loa}
\providecommand{\algorithmname}{Algorithm}
\floatname{algorithm}{\protect\algorithmname}

\makeatletter
\let\oldforeign@language\foreign@language
\DeclareRobustCommand{\foreign@language}[1]{%
	\lowercase{\oldforeign@language{#1}}}

\DeclareTextSymbolDefault{\textquotedbl}{T1}

\let\oldforeign@language\foreign@language
\DeclareRobustCommand{\foreign@language}[1]{%
	\lowercase{\oldforeign@language{#1}}}

\ifCLASSINFOpdf
\else
\fi

\@ifundefined{showcaptionsetup}{}{%
	\PassOptionsToPackage{caption=false}{subfig}}
\usepackage{subfig}

\newcommand{\MYfooter}{\smash{
		\hfil\parbox[t][\height][t]{\textwidth}{\centering
			\thepage}\hfil\hbox{}}}

\def\ps@IEEEtitlepagestyle{%
	\def\@oddhead{\parbox[t][\height][t]{\textwidth}{\centering \scriptsize
			Personal use of this material is permitted. Permission from the author(s) and/or copyright holder(s), must be obtained for all other uses. Please contact us and provide details if you believe this document breaches copyrights.\\
			\noindent\makebox[\linewidth]{}
		}\hfil\hbox{}}%
	\def\@evenhead{\scriptsize\thepage \hfil \leftmark\mbox{}}%
	\def\@oddfoot{\parbox[t][\height][l]{\textwidth}{
			\vspace{-20pt}{\rule{\textwidth}{0.4pt}}\\ \footnotesize\underline{To cite this article:}
			{\bf{\footnotesize\textcolor{red}{A. V. Jonnalagadd and H. A. Hashim, "SegNet: A Segmented Deep Learning based Convolutional Neural Network Approach for Drones Wildfire Detection," Remote Sensing Applications: Society and Environment, 2024.}}} doi: \href{https://doi.org/10.1016/j.rsase.2024.101181}{10.1016/j.rsase.2024.101181}\\
			\noindent\makebox[\linewidth]
		}\hfil\hbox{}}%
	\def\@evenfoot{\MYfooter}}

\makeatother
\begin{document}
	\bstctlcite{IEEEexample:BSTcontrol}

\title{SegNet: A Segmented Deep Learning based Convolutional Neural Network Approach for Drones Wildfire Detection}

	\author{Aditya V. Jonnalagadda and Hashim A. Hashim
		\thanks{This work was supported in part by National Sciences and Engineering
			Research Council of Canada (NSERC), under the grant number RGPIN-2022-04937.}
		\thanks{A. V. Jonnalagadda and H. A. Hashim are with the Department of Mechanical
			and Aerospace Engineering, Carleton University, Ottawa, ON, K1S-5B6,
			Canada (e-mail: AdityaVardhanJonnal@cmail.carleton.ca and hhashim@carleton.ca).}
	}

	
	
	\maketitle
	\begin{abstract}
		This research addresses the pressing challenge of enhancing processing
		times and detection capabilities in Unmanned Aerial Vehicle (UAV)/drone
		imagery for global wildfire detection, despite limited datasets. Proposing
		a Segmented Neural Network (SegNet) selection approach, we focus on
		reducing feature maps to boost both time resolution and accuracy significantly
		advancing processing speeds and accuracy in real-time wildfire detection.
		This paper contributes to increased processing speeds enabling real-time
		detection capabilities for wildfire, increased detection accuracy
		of wildfire, and improved detection capabilities of early wildfire,
		through proposing a new direction for image classification of amorphous
		objects like fire, water, smoke, etc. Employing Convolutional Neural
		Networks (CNNs) for image classification, emphasizing on the reduction
		of irrelevant features vital for deep learning processes, especially
		in live feed data for fire detection. Amidst the complexity of live
		feed data in fire detection, our study emphasizes on image feed, highlighting
		the urgency to enhance real-time processing. Our proposed algorithm
		combats feature overload through segmentation, addressing challenges
		arising from diverse features like objects, colors, and textures.
		Notably, a delicate balance of feature map size and dataset adequacy
		is pivotal. Several research papers use smaller image sizes, compromising
		feature richness which necessitating a new approach. We illuminate
		the critical role of pixel density in retaining essential details,
		especially for early wildfire detection. By carefully selecting number
		of filters during training, we underscore the significance of higher
		pixel density for proper feature selection. The proposed SegNet approach is rigorously evaluated using real-world
		dataset obtained by a drone flight and compared to state-of-the-art
		literature.
	\end{abstract}
	

	\IEEEpeerreviewmaketitle{}
	\rule{0.47\textwidth}{1pt}\\
	Video of the experiment: \href{https://youtu.be/xwMzFpZkC8M}{Click Here}\\
	\rule{0.49\textwidth}{1pt}
	
	\section{Introduction}
\begin{figure*}
		\centering{}\includegraphics[scale=0.26]{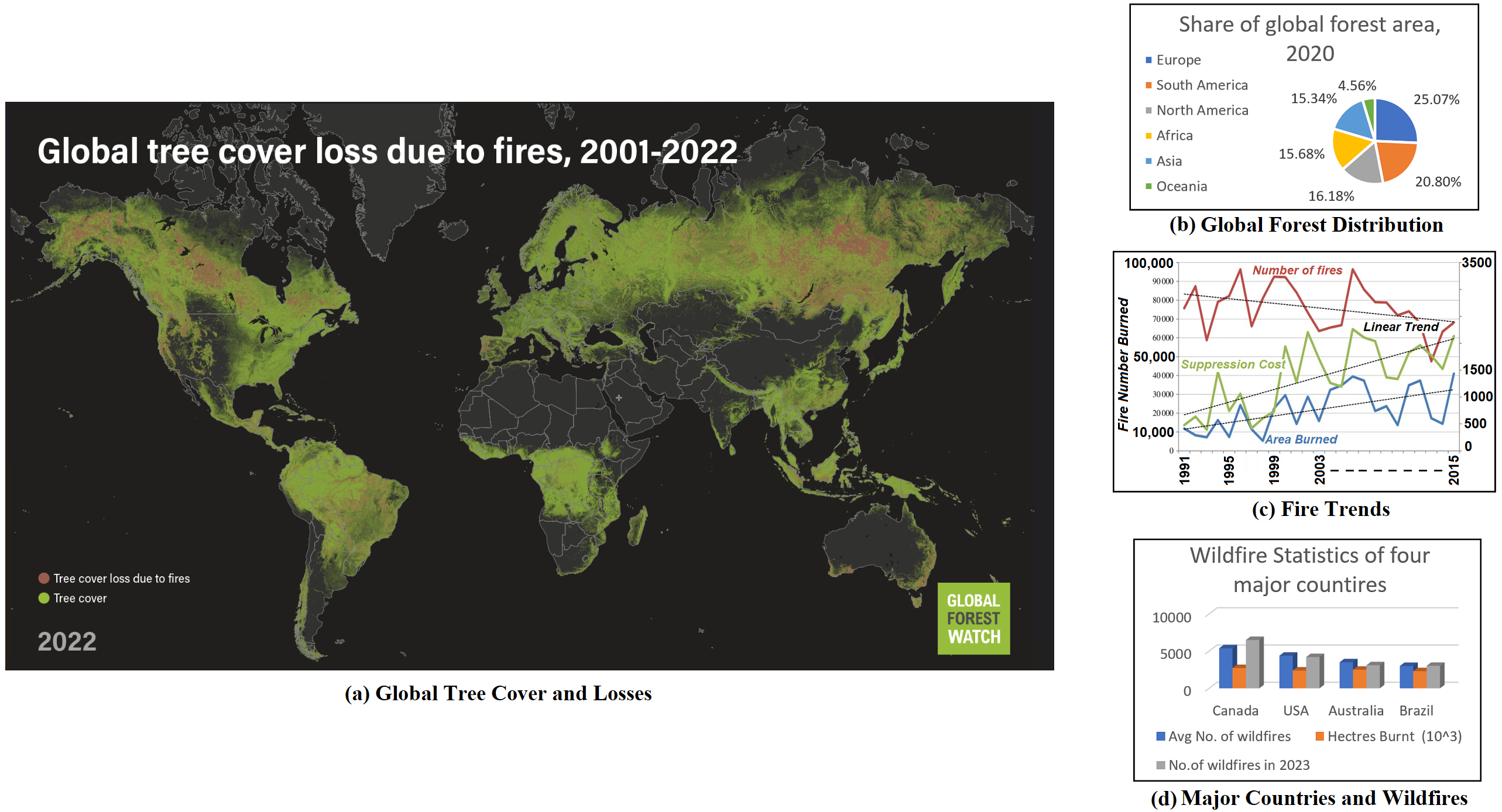}\caption{Wildfires is a global and growing threat: (a) Global tree cover loss
			due to forest fires, 2001-2022 (courtesy to \cite{Ref32}). (b) Global
			distribution of forest land \cite{Ref44}. (c) Number of fires and
			suppression cost with area burnt along with linear trend lines from
			1991-2015 (National Interagency Fire Center \cite{Ref42}). (d) Wildfire
			metrics of Canada, USA, Australia, and Brazil \cite{Ref27}.}
		\label{fig:Fig1}
	\end{figure*}

	\IEEEPARstart{O}{ur} lives rely heavily on the resources that forests provide. They
	are regarded as the planet’s lungs because they filter the air by
	adding oxygen (O2) and lowering the high levels of carbon dioxide
	levels (CO2). They serve as homes for a variety of animals and can
	be utilized to shield crops from the wind. Additionally, they clear
	the water of the majority of pollution-causing agents \cite{Ref39,Ref43}.
	Due to the numerous jobs and higher revenues that forests create,
	countries’ economies are improved. Forests have a profound impact
	on humanity by providing essential ecosystem services. They purify
	air, regulate climate, protect against natural disasters, and support
	biodiversity. Additionally, forests offer resources like timber and
	medicines, while promoting recreation and cultural heritage, highlighting
	their critical role in sustaining human well-being and the planet.
	Forest fires, often exacerbated by factors like climate change and
	human activity, have devastating effects on ecosystems, communities,
	and the environment. Forest fires, raging with increasing frequency
	and intensity, inflict profound damage. Ecologically, they destroy
	vital habitats, decimate wildlife populations, and disrupt ecosystems.
	Native flora and fauna struggle to recover, and invasive species often
	take hold in the aftermath. Fig. \ref{fig:Fig1}.(a) shows the global
	tree cover loss occurred between the years 2001 to 2022 \cite{Ref32}.
	Red dots are the areas effected by forest fires, few of which are
	under serious efforts of restoration. Communities near forests face
	immediate peril, with lives and homes in jeopardy. Firefighters risk
	their lives battling infernos. Smoke and air pollution pose serious
	health threats, especially to vulnerable populations. Evacuations
	disrupt livelihoods and cause psychological trauma. Economically,
	the costs are staggering. Firefighting expenditures soar, and losses
	in timber, agriculture, and tourism industries mount. Long-term, diminished
	soil fertility hinders agriculture, and reduced water quality impacts
	communities downstream. Environmental repercussions extend globally.
	Forest fires release vast amounts of carbon dioxide, contributing
	to climate change. This, in turn, exacerbates conditions conducive
	to more frequent and severe fires in a vicious cycle. Fig. \ref{fig:Fig1}.(b)
	shows the share of total global forest area across continents \cite{Ref44}.
	
	\subsection{Related Work}
	
	Preventing and mitigating forest fires requires concerted efforts.
	Strategies include controlled burns, firebreaks, and early warning
	systems. Additionally, addressing climate change and promoting sustainable
	land management are crucial to curbing the catastrophic effects of
	these infernos. In recent years, a lot of forest fires have been taking
	place and because of wildfire’s terrifying effects on the economy,
	human health, and the environment, forest accidents have emerged as
	one of the greatest risks to humanity. Every year, more than $8,000$
	fires burn an average of more than 2.1 million hectares in only Canada
	\cite{Ref2}. Wildfire frequency and total burnt area alone in the
	western US have surged by $400\%$ during the past ten years \cite{Ref34}.
	The wildfire in Australia was the most devastating fire in 2020, resulting
	in numerous losses, including the destruction of more than $1,500$
	homes, the death of around $500,000$ animals, and the burning of
	more than 14 million acres \cite{Ref1}. Other destructive wildfires
	have occurred, causing enormous losses. For example, the 2018 California
	fire and the 2019 Amazon rain forest fire both burned millions of
	acres of land. Based on data in the National Forestry Database, over
	$8,000$ fires occur each year, and burn an average of over 2.1 million
	hectares in Canada \cite{Ref2}. Looking at the linear trends in Fig.
	\ref{fig:Fig1}.(c), the inclining slope of suppression costs depict
	an ever-growing effort to curb forest fires \cite{Ref42}. Although
	the steep inclining slope of burnt area does not justify the slow
	declining rate in number of fires despite the technological advances
	(cost slope). As proved by the wildfire cases stated earlier and the
	statistics from Fig. \ref{fig:Fig1}.(d) wildfire occurrences have
	caused increased damage every passing year throughout the decade.
	Canada, USA, Australia, and Brazil \cite{Ref27} are amongst the top
	hotspots for frequent wildfires. Accordingly, from Fig. \ref{fig:Fig1}
	it is notable that overall wildfires are in a growing uptrend and
	novel solutions are of great importance and urgent demand.
	
	Sensors and sensor fusion are standard part for early prediction/detection and therefore risk mitigation and management \cite{hashim2023uwb,hashim2021geometric,hashim2020landmark}. Typically, sensors like gas, smoke, temperature, and flame detectors
	are used to identify wildfires \cite{Ref14,Ref7,Ref15,liu2023remote,bushnaq2021role}.
	However, these detectors have a number of drawbacks, including slow
	response times and limited coverage areas \cite{Ref14}. Traditional
	fire detection methods are now being superseded by vision-based models
	because of their accuracy, wide coverage regions, low error probability.
	Moreover, vision-based techniques are distinguished with compatibility
	with current camera surveillance systems which can be easily implemented
	\cite{hashim2023exponentially,hashim2023observer}. In order to create
	precise fire detection systems, researchers have proposed numerous
	cutting-edge computer vision approaches over the last years, including
	Infra-Red (IR) images \cite{hashim2023observer,Ref37}. As a result,
	Unmanned Aerial Vehicle (UAV) or drone systems have recently gained
	popularity and been used to combat and find the forest’s deep wildfires.
	Additionally, by combining this with Deep Learning (DL) methods, it
	has made remarkable progress \cite{Ref4}. The color of the wildfire
	and its geometrical characteristics, such as angle, shape, height,
	and width, are detected using Deep Learning (DL)-based fire detection
	algorithms. Their findings are fed into models of fire propagation
	\cite{Ref4,Ref5}. The effectiveness of these approaches for identifying
	and segmenting forest fires from UAV photos, in a real-life case,
	is yet unknown and to-date a challenging open problem. In particular,
	in the light of several difficulties that must be addressed, such
	as the tiny object size, the complexity of the backdrop, and the image
	deterioration and high computational cost \cite{Ref16,Ref17}. 
	
	Several research efforts have been established concerning forest wildfire
	detection and prediction \cite{Ref20,Ref23,Ref21}. The research focus
	and implementation range from fixed to mobile solutions. Although
	mobile solutions have gained the support of many researchers as it
	outweighs its counterpart considering their advantages. Static solutions
	often deal with watchtowers that can be expensive to install \cite{Ref20}
	and carry the risk of being damaged by the fire itself, adding unnecessary
	repair costs on top. They are also restricted by the field of vision
	and obstruction of vision by complete or partial forest canopies.
	Mobile solutions such as UAVs can provide data in the form of images
	from different angles considering their flexibility to maneuver in
	the six degrees-of-freedom (6 DoF) \cite{Ref23}. Nowadays, several
	researchers consider a similar approach to solving this problem which
	go through the following steps: (1) Data Acquisition System: acquiring
	data using UAV sensors and cameras; (2) Data Processing Onboard: processing
	data with onboard microchips with learning algorithms; (3) Data Transmission/Receiving
	System: data is either transmitted to or received by on-ground equipment
	for further studies of data; and (4) Notifying Concerned Authorities:
	the wildfire management authorities are informed to make accommodations
	for further actions required \cite{Ref21}.
	
	\subsection{Persisting Challenges}
	
	For any real-time wildfire detection algorithm both the space and
	time resolutions are important. Stationary imagery has good time resolution
	but low spatial resolution. This is because the satellite body or
	watch tower can house a heavy, high performing GPU to process faster
	decreasing the processing times. Although, dispersion of light at
	various different angles, information in each pixel and environmental
	factors like fog, cloud cover, etc, hinder satellite’s vision capabilities
	resulting in low spatial resolution. On the contrary, mobile imagery
	consists of good spatial resolution but has poor time resolution.
	Thus, drone or rover imagery with conventional fire detection methods
	(smoke sensors, temperature sensors, etc) are not practically effective
	in real-time scenarios. Limitations in this area of research are plenty,
	these prove to be an obstacle for any researcher to fluently conduct
	their study. One such case would be the selection of Graphics Processing
	Units (GPU) since computational burden is a key element. High powered
	GPUs are not feasible as they have higher mass and volume, which are
	not preferred due to the limited weight capacity of drones. One other
	such limitation that is often overlooked is the problem of detecting
	forest fires during the fall season. Sugar maples in Canada or trees
	from the Laurel family around the globe turn orange-red in the fall
	season rendering any pixel color-based image classification technique
	ineffective, increasing the number of false alarms.
	
	In case of any DL network, density of network is directly proportional
	to the delay caused in perceiving an image and classifying the image
	as fire or non-fire. Generally, a denser neural network increases
	the accuracy of the predictions due to increased processing of image
	feature pool. Although, for a real-time application, the algorithm
	must contain fewer dense layers for faster processing speeds and this
	must be carried out without having to compromise on the accuracy.
	Limited dataset of wildfire images available is one such gap that
	is slowing down the development of wildfire detection. With scarce
	resources of images available on the internet and real-life drone
	footage, capturing the aerial shot of wildfire, requiring government
	permissions to shoot and use, researchers have turned their focus
	on either augmenting the dataset or increasing the efficiency by using
	new enhanced algorithms. Datasets are being appended with newer synthetic
	images using Generative Adversarial Networks (GANs) \cite{Ref35},
	where a real non-fire image is translated into a modified image with
	fire. This also helps with creating wildfire datasets for YOLOv3 \cite{Ref9}
	format with localization and bounding box coordinates of fire within
	the image. Frame to frame capture of images from a wildfire video
	can produce large number of images but these lack in diversity of
	data which could cause overfitting to occur. Flames and smokes are
	amorphous in nature and difficult to label. Thus, to expertly annotate
	each image can take huge amounts of time. Therefore, the use of algorithms
	such as YOLOv3 \cite{Ref9} or faster Region-based Convolutional Neural
	Network (R-CNN) \cite{Ref41} becomes limited due to the lack of labeled
	and formatted data for training and testing of these algorithms.
	
	\subsection{Modern Machine Learning Approaches}
	
	DL techniques have been popular over the past few decades as alternate
	strategies for difficult issues, such as the management and forecast
	of wildfires. The work in \cite{Ref5} reported a comparison research
	using Artificial Neural Networks (ANN) and Local Regression (LR) to
	map fire susceptibility, coming to the conclusion that ANN outperformed
	LR. A comparison of a deterministic approach and two Machine Learning
	(ML) techniques-Radio Frequency (RF) and extreme learning machine
	was provided by \cite{nakao2022assessing,Ref42,Ref6,charizanos2023bayesian}.
	The results of this investigation showed that the three techniques
	performed equally, emphasizing the advantage of both stochastic methods
	in that they are data-driven and, as a result, independent of prior
	information. Support Vector Machine (SVM), RF, and Multi-layer Perceptron
	(MLP), three ML-based approaches, were evaluated in a further comparison
	research by \cite{thach2018spatial}, with the MLP achieving the highest
	accuracy score \cite{Ref8}. Three techniques were employed by \cite{Ref36}
	for multi-hazard modeling (namely snow avalanches, floods, wildfires,
	landslides and land subsidence). Generalized Linear Model (GLM), SVM,
	and functional discriminant analysis were the approaches used. GLM
	produced the best results for predicting the risk of wildfires, closely
	followed by the other two. The work in \cite{Ref13} mapped fire susceptibility
	using the General Additive Model (GAM), Multivariate Adaptive Regression
	Spline (MARS), SVM, and the ensemble GAM-MARS-SVM. SVM was shown to
	be the least accurate approach out of these, whereas the ensemble
	had the best predictive accuracy. Most of the research is dealt with
	Convolution Neural Networks (CNN) and its variants, Single Shot Multi-Box
	Detector (SSD), U-shaped encoder-decoder network (U-Net), and deep
	Lab \cite{Ref9}. 
	
	Some other less popular but effective learning algorithms include
	Long Short-Term Memory (LSTM), Deep Belief Network, and Generative
	Adversarial Network \cite{Ref35}. These algorithms are not preferred
	as they require powerful hardware which is difficult to house in a
	mobile UAV. The additional pieces of hardware add to the weight of
	the UAV creating a problem in terms of flight and control \cite{Ref31}.
	The use of remote compact cameras on UAVs outweighs the performance
	of the images taken by satellites due to their capacity to capture
	higher pixel density images from different angles. A tiny fire spot
	or extreme dryness of the objects would be almost impossible to spot
	using the images by satellites. Hence, to train the model better and
	to yield better results, mobile cameras are used. Additionally, the
	significance of the dataset has been consistently stressed to enhance
	the performance of the model since neural networks cannot be applied
	to untrained scenarios. Because it is difficult to detect smoke during
	the night and because the color and texture of smoke during model
	verification are too similar to other natural phenomena like fog,
	clouds, and water vapor, algorithms that rely on smoke detection typically
	have issues like high false alarm rates \cite{Ref40}.
	
	\subsection{Contributions}
	
	In this paper we address the problem of time resolution for UAV drone
	imagery along with limited dataset available on wildfires around the
	world using a proposed Segmented Neural Network (SegNet) selection
	approach based on DL reducing the total feature map. This novel SegNet
	technique contributes the following to robotic wildfire surveillance
	systems: (a) Increased processing speeds enabling real-time detection
	capabilities for wildfire in shorter period of time when compared
	to the existing cutting-edge solutions (e.g., \cite{Ref38,Ref10});
	(b) Increased detection accuracy of wildfire which is confirmed through
	training, testing, and validation; (c) Introducing a new direction
	for image classification of amorphous objects which can add significant
	insight to fire, water, smoke, etc; and (d) Improved detection capabilities
	which can be potentially employed for early wildfire detection.
	
	\subsection{Structure}
	
	The rest of the paper is organized as follows: Section \ref{sec:Problem-Formulation}
	problem formulation, dataset preparation, data augmentation, and scaling
	issues. Section \ref{sec:Methodology} illustrate the research methodology
	and segmentation. Section \ref{sec:WorkFlow} presents workflow, challenges,
	and mitigation. Section \ref{sec:Results} illustrate results of the
	proposed SegNet approach in comparison to state-of-the-art literature.
	Finally, Section \ref{sec:Conclusion} summarizes the work.
	
	\section{Problem Formulation\label{sec:Problem-Formulation}}
	
	The rapid advancements in Artificial Intelligence (AI) have catapulted
	it into one of the most swiftly evolving fields in applied science.
	However, amidst this complexity, researchers striving to emulate the
	intricacies of the human brain have crafted algorithms so sophisticated
	that they present unique challenges, particularly in their integration
	into the engineering sector. In the realm of robotics engineering,
	these challenges become evident. One of the significant hurdles stems
	from the complexity of machine learning models, which demand powerful
	computational processing. The current technological landscape grapples
	with a limitation: the lack of powerful, robust yet lightweight processing
	systems that can be seamlessly integrated into mobile robots. Considering
	drones or rovers, for instance. Equipping them with hefty, powerful
	processors is unfeasible, as these components are often heavy and
	can hamper the mobility and energy efficiency of these autonomous
	devices. Moreover, the cost factor amplifies the issue; these potent
	processing systems are expensive to produce, rendering them economically
	unviable for mass production. This becomes a critical concern, especially
	in the context of wildfire detection across extensive land masses.
	The application of complex machine learning algorithms on relatively
	weaker processors exacerbates the problem. While using less powerful
	processors in an attempt to mitigate the weight issue, the computational
	time required for running intricate algorithms becomes substantial,
	rendering these processors ineffective in real-time applications.
	Consequently, there arises a pressing need for a novel approach in
	machine learning---one that can provide real-time decision-making
	capabilities in the realm of wildfire detection. This necessity is
	steering researchers towards innovative solutions, emphasizing the
	urgency to bridge the gap between the robustness of algorithms and
	the limitations of current processing technologies in the pursuit
	of efficient and timely wildfire detection systems.
	
	\subsection{Dataset and Preparation}
	
\begin{figure}
		\centering{}\includegraphics[scale=0.22]{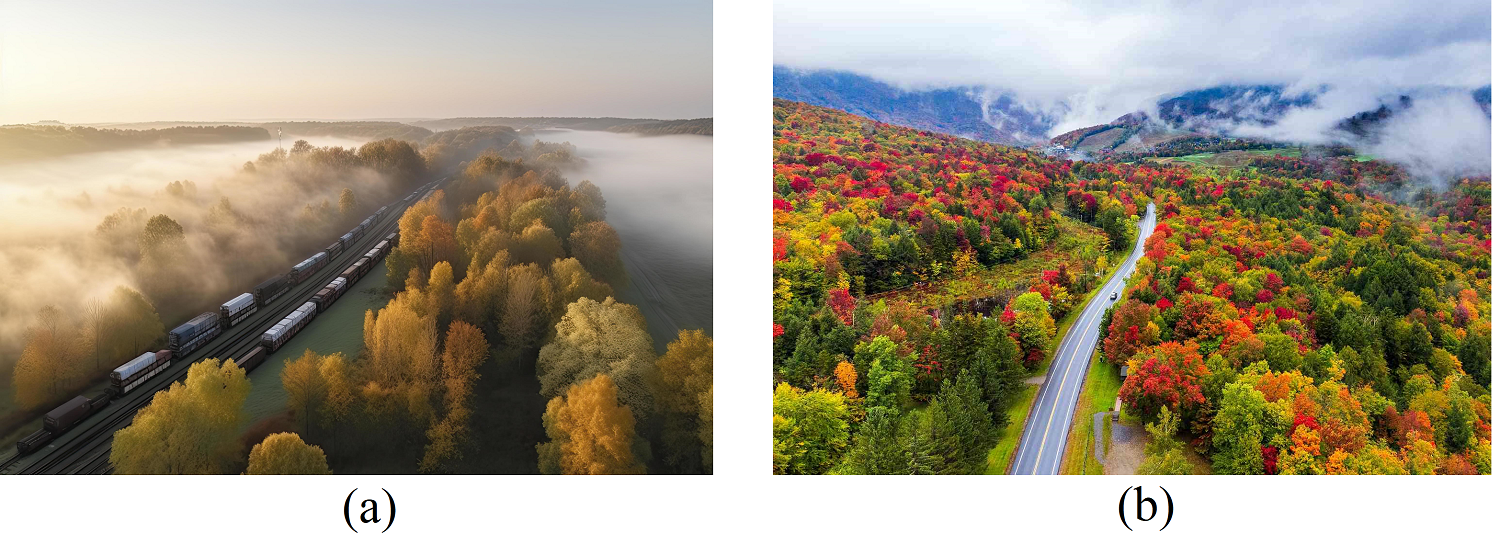}\caption{Environmental challenges: (a) Images with fog can be misclassified
			as smoke caused by wildfire, and (b) Red trees in the forest could
			yield undesired false results by model due to misclassification of
			the red spots (trees) in the image \cite{Ref50}.}
		\label{fig:Fig2}
	\end{figure}

	Some images in the dataset might prove to counter the ideology being
	pursued in this research paper, that is, to subject the model to select
	features of fire like smoke, amorphous shape of the fire, color and
	brightness, etc (see Fig. \ref{fig:Fig2}). Some images contain environmental
	factors like fog and mist and can be misrepresented by the model such
	as smoke caused by fire. This eventually yields false positives during
	testing of the model. Some images that were acquired in the fall season
	often contain trees that have the fall coloring, that is, they turn
	into hues of orange, red and maroon. These hues of colors are similar
	to the colors of flame which are challenging. Thus, the model might,
	when trained with fall season images, classify an image with fire
	as non-fire attributing the fire image to be a fall season colored
	tree.
	
	\subsection{Segmentation vs Complete Image}
	
	\begin{figure}
		\centering{}\subfloat[Complete image.]{\includegraphics[scale=0.62]{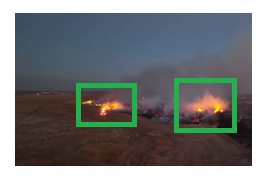}}
		\subfloat[Segmented image.]{\includegraphics[scale=0.59]{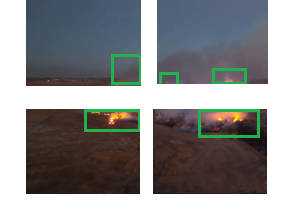}}
		
		\subfloat[Normal image.]{\includegraphics[scale=0.125]{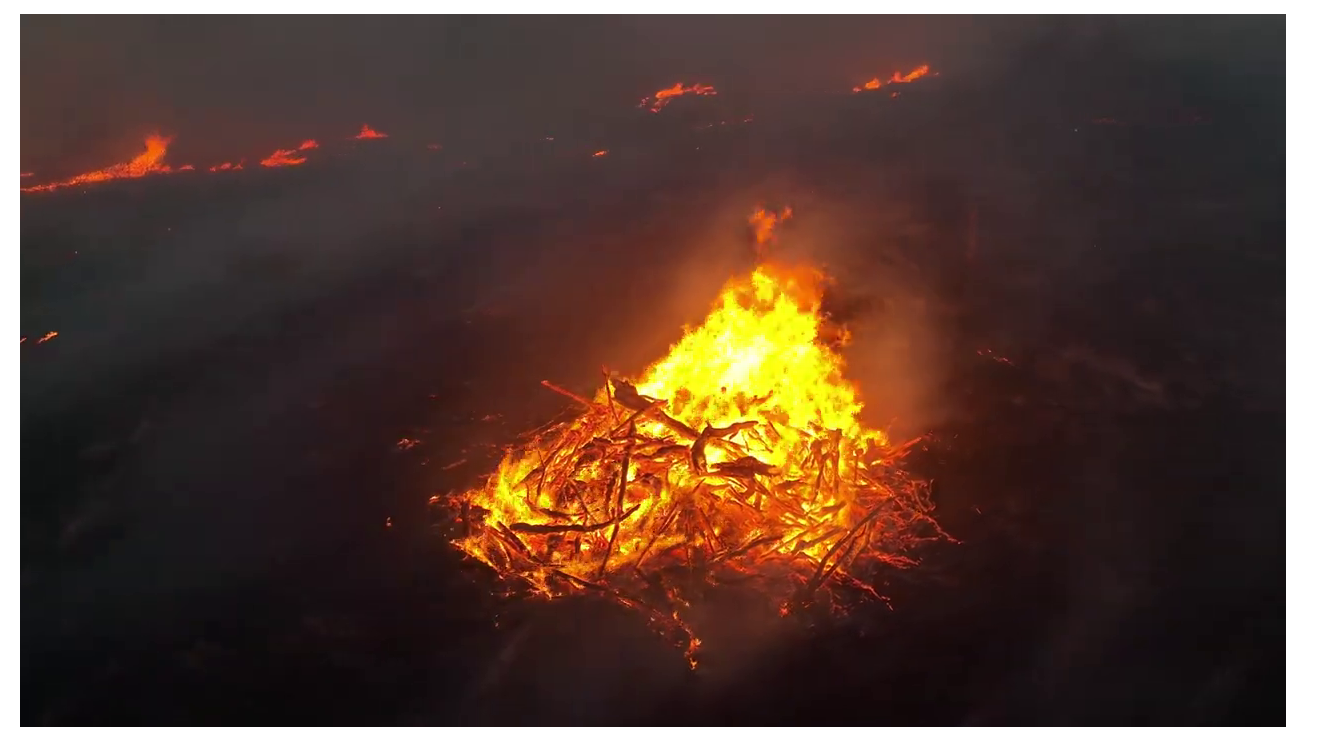}} \subfloat[Augmented image.]{\includegraphics[scale=0.125]{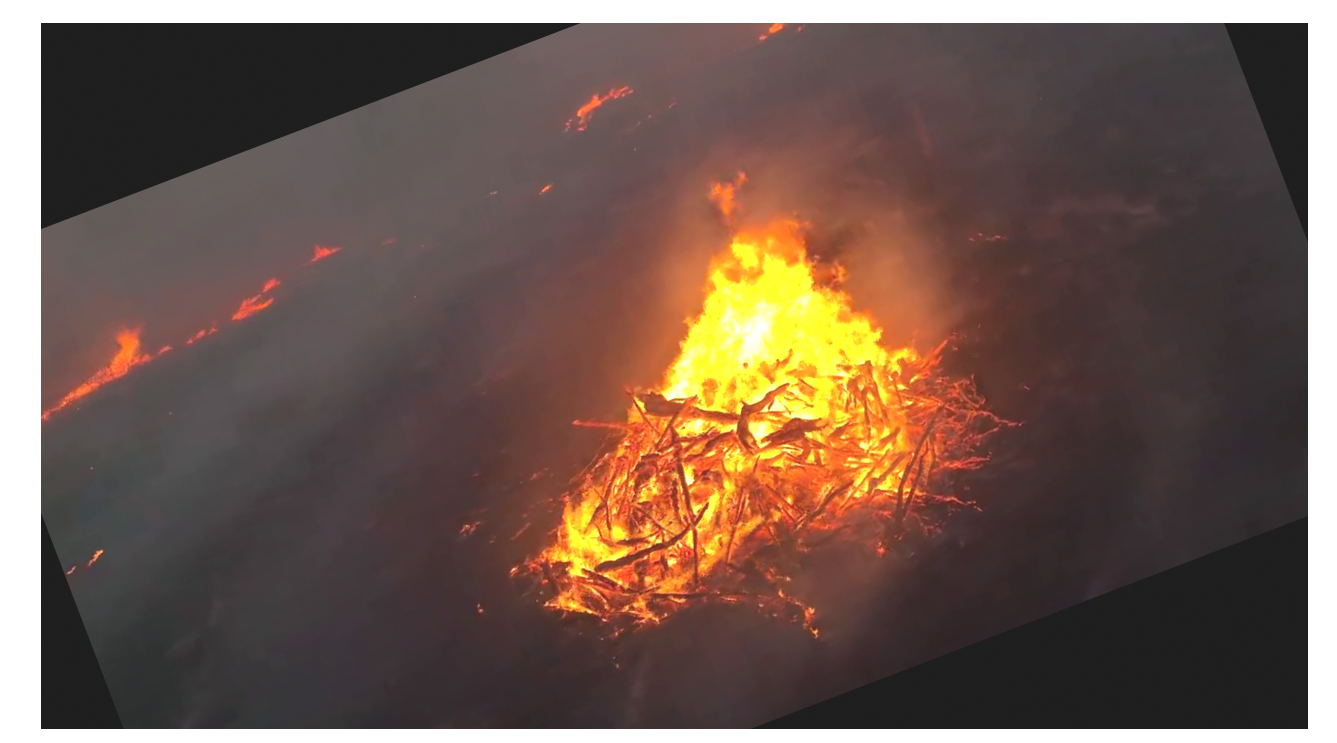}}
		
		\subfloat[No scaling.]{\includegraphics[scale=0.48]{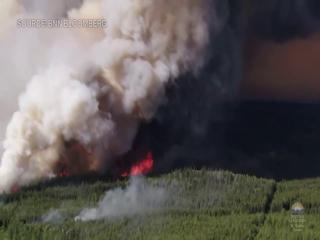}}\subfloat[Rescaling introduced.]{\includegraphics[scale=1.7]{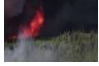}}\caption{Dataset and preparation: (a) flames in a complete image, (b) flames
			in a segmented image, (c) flames in normal image (courtesy to WXChasing),
			and (d) flames in an augmented image, (e) no scaling (courtesy to
			Bloomberg), (f) Loss of information when scaling an image’s resolution.}
		\label{fig:Fig3}
	\end{figure}
	
	Data for this classification problem is often subjected to limitations
	in the form of availability and proper labeling of classes. It is
	often stated by researchers \cite{Ref22}, that lack of proper dataset
	for wildfires, unlike some other applications that include weather
	forecast, stock market predictions, models built around NLP like chatbots
	and smart AI assistants, has always limited the approach towards addressing
	this problem using various different state of the art techniques.
	Techniques like YOLOv3, R-CNN require training datasets with predefined
	bounding boxes around the class being investigated. Faster R-CNN as
	stated by \cite{Ref41} has proven to yield faster speeds in processing
	with high accuracy although the variability of training dataset is
	difficult to achieve due to limited dataset with bounding boxes available.
	Thus, in this research a custom dataset is used to train the model
	comprising of both complete images and segmented images (visit Fig.
	\ref{fig:Fig3}.(a) and \ref{fig:Fig3}.(b)). Each complete image
	can be broken down into 12 different segmented images. It is important
	to downsize the complete image from a resolution of $1280\times720$
	pixels to $320\times240$ pixels. This is performed by using the PILLOW
	library available in the python environment. The combination of complete
	images and segmented images allow for the model to look for features
	of flame as a complete and incomplete demonstration.
	
	\subsection{Data Augmentation}

	To improve dataset accuracy, various techniques were used to create
	diverse data. The initial model showed high accuracy in training but
	struggled during validation, indicating overfitting. To tackle this,
	artificial images were made through augmentation, preventing the model
	from focusing too much on specific examples. This helped the model
	recognize broader patterns and reduced the risk of memorizing isolated
	instances. The augmentation also addressed real-world challenges like
	different lighting conditions and viewpoints, common in practical
	scenarios throughout the year. Techniques including rotation, translation,
	scaling, brightness adjustment, and the introduction of Gaussian noise
	were tactically applied \cite{Ref33,Ref18}. These methods collectively
	fortified the model’s adaptability, enabling it to navigate through
	various input scenarios with resilience and precision (visit Fig.
	\ref{fig:Fig3}.(c) and \ref{fig:Fig3}.(d)). By incorporating these
	augmentation strategies, the model has not only become adept at handling
	diverse and nuanced data but also emerged as a robust tool for subsequent
	analyses and predictions. The dataset, enriched through these interventions,
	provided a solid foundation for the model to generalize patterns effectively,
	ensuring its applicability in real-world situations. This meticulous
	approach not only elevated the model’s training accuracy but also
	its validation accuracy.
	
	\subsection{Scaling Issues}
	
	In Fig. \ref{fig:Fig3}.(e) and \ref{fig:Fig3}.(f), two different
	pixel sizes of the same shot of wildfire are taken. An image with
	$320\times240$ pixels (on the left) and an image with $1280\times720$
	pixels (on the right). Same segments of the both these images are
	extracted which have the same area covered. The image with higher
	pixel density exhibits sharper image contrary to the image with lower
	pixel density which exhibits loss of information. This can be observed
	by looking at the trees in the environment. The shape and texture
	are dull when a lower pixel size is considered. This loss of information
	might result in improper filter selection during training effecting
	the accuracy of the model. Early detection of wildfire contains detection
	of a small fire spot and thus having higher pixel density is important
	and necessary.
	
	\section{Methodology\label{sec:Methodology}}
	
	CNNs have always been on the forefront when it comes to image classification
	problems. This is mainly due to the fact that CNN works with the use
	of several different filters to extract features out of an image \cite{Ref25}.
	Feature extraction and engineering is a domain in deep learning processes
	that has gained specific interest of lately. Yet, there is still a
	lot of development and understanding required about the same. This
	paper focuses on introducing a new practice of feature reduction in
	an image. In fire detection algorithms developed for data in the form
	of live feed data focuses on several features that cannot be assessed
	in an image classification algorithm. The constant flickering and
	moving pixels in the video data presents a quick solution for detection.
	Although this method falls behind in time resolution. In this paper
	we focus on image feed rather than video feed. The field of image
	classification has seen a lot of development towards improving the
	accuracy of detection. Very little focus has been put towards improving
	processing times to make real time detection effective. This algorithm
	focuses on reducing the number of features in the input image. This
	is done by segmentation of image. Any image feed when fed into the
	neural network, especially CNN, contains numerous features. These
	features could be anything ranging from objects and environment. These
	can be further divided into distinctions such as shape of the object,
	color of the object, texture of the object or how the object interacts
	with the environment. When in neural network convolutions are carried
	out by the filters, the product received is known as a feature. The
	collection of such features produced by various filters is known as
	a feature map. Higher number of image pixels translate to richer information
	about the image. More information then translates to higher number
	of total relevant features in the image.
	\[
	\text{Number of pixels in image}\propto\text{information in the image}
	\]
	\[
	\text{Image information}\propto\text{Number of discrete features}
	\]
	For every engineering problem, choice of number of features in feature
	map changes. Some applications like anomaly detection in mechanical
	parts in a factory requires high pool of feature map to properly distinguish
	between different types of manufacturing defects. Similarly for wildfire
	detection problem a larger feature map enables capturing of the fire
	and its characteristics but negatively effects the training stage.
	A larger feature map requires a larger dataset with highly varying
	images of fire and non-fire which is neither easily available nor
	can they be effectively procured. An insufficient dataset while training
	the algorithm with larger feature map ends up with underfitting of
	the model. Data augmentation techniques also do not provide any better
	results as it leads to overfitting of the data. This is one of the
	main reasons for researchers using images of $256\times256$ image
	size which results in significantly smaller feature map. This also
	results in loss of information.
	
	\subsection{Segmentation}
	
	\begin{figure*}
		\centering{}\includegraphics[scale=0.26]{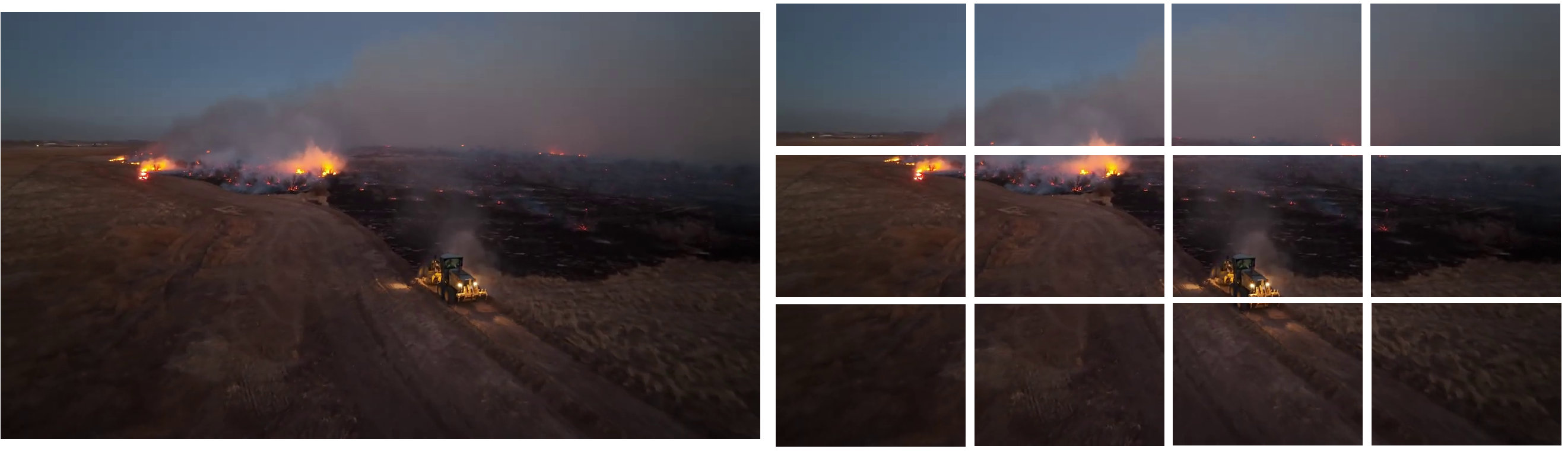}\caption{Segmentation of a complete image into twelve segmented images avoids
			loss of critical information and enables fire spots to be the prominent
			feature in the segmented image for better feature detection and selection.
			This approach can only be implemented on objects, under detection,
			that are amorphous in nature (do not exhibit a consistent shape or
			form).}
		\label{fig:Fig4}
	\end{figure*}
	
\begin{figure*}
		\centering{}\includegraphics[scale=0.45]{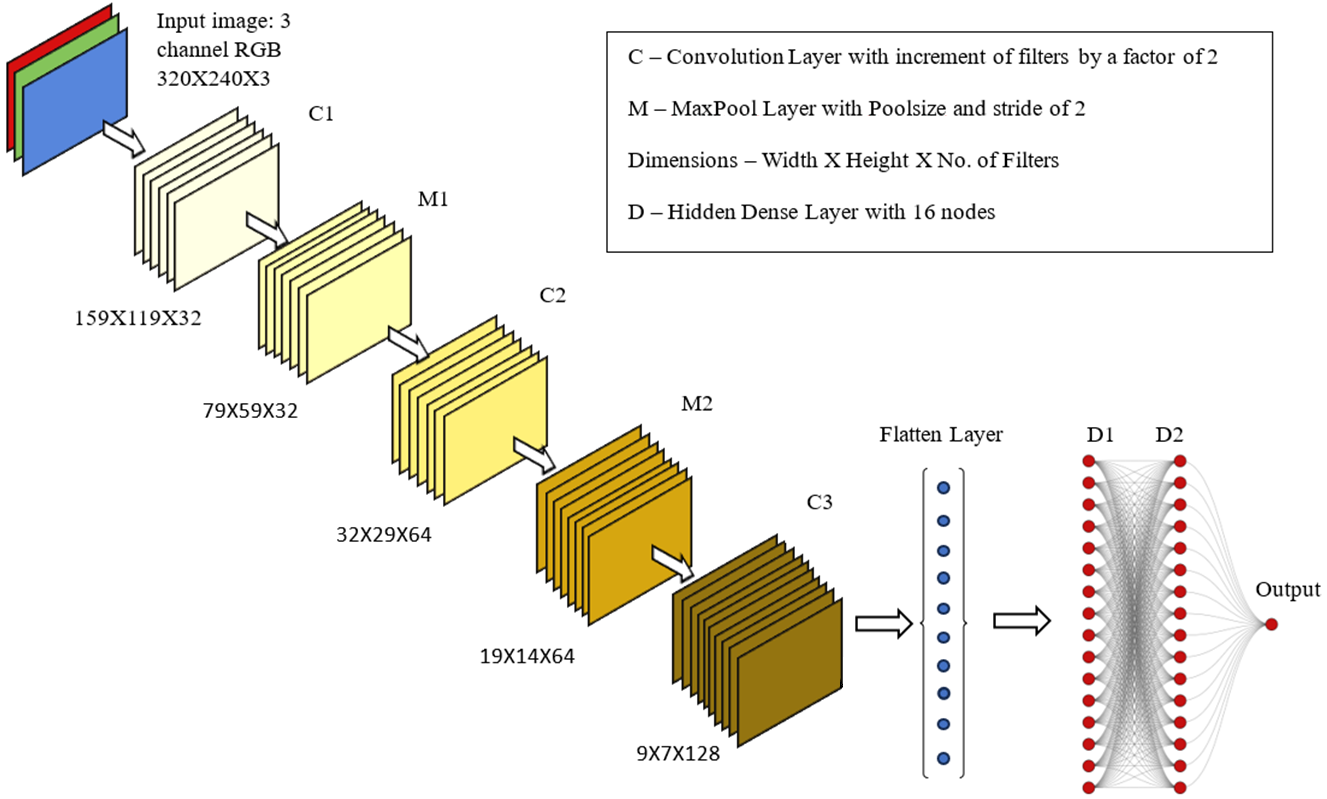}\caption{The proposed SegNet architecture specifically designed to be light
			on computations. This model can be deployed in drones with weak processing
			systems. The input image to this architecture is a segmented image
			achieved after segmentation of a complete $1280\times720$ image into
			12 smaller $320\times240$ segments.}
		\label{fig:Fig5}
	\end{figure*}

	Thus, in this paper we propose a solution using the segmentation technique
	of a wildfire image. In this technique, a high pixel image is broken
	down into several pieces. These pieces are equal in pixel size and
	are considered as different images. When any image is cut into pieces,
	that can be concatenated to form the original image, information is
	not lost. It is important to note that several objects in the image
	may lose their integrity in terms of their shape. For example, in
	Fig. \ref{fig:Fig4}, a tractor has lost its integrity of shape as
	it is segregated into two different images (body is in one image and
	the tires are in the other). This could hinder the algorithm’s capability
	to detect that object. Classification problems that involve morphous
	object’s detection, for an instance, an image of a person, are hard
	to deal with due to the same aforementioned reason. Thus, this technique
	is not useful for detection of fixed shaped objects. Flames, on the
	contrary, are amorphous in nature, that is, they do not possess a
	fixed shape or form. They can be found in different shapes, forms,
	color and intensity. This amorphous nature of fire enables this segmented
	image approach of this paper’s algorithm. Number of segments to be
	made of an image is dependent on the pixel size of the original image
	and the required processing speeds. Smaller the pixel size of segment
	of an image, faster is the processing speed involved. In this paper,
	original image pixel size of $1280\times720$ is used in the dataset.
	These images are segmented into 12 segments where each segment is
	$320\times240$ pixels pieces.
	
	\[
	\frac{\text{Width of original image}}{\text{Number of vertical segments}}=\frac{1280}{4}=320\text{ pixels}
	\]
	\[
	\frac{\text{Height of original image}}{\text{Number of horizontal segments}}=\frac{720}{3}=240\text{ pixels}
	\]
	\[
	\text{Total number of segmented images}=4\times3=12
	\]
	Segmentation of images into smaller pieces cause two positive outcomes.
	First, in early wildfire images, flames become a prominent feature
	of one of the segmented-images. This, makes it easier for algorithm
	to detect flames and smoke resulting in increased accuracy for early
	wildfire detection. Second, processing speeds for these segmented-images
	will now increase as number of pixels of input image decreases (less
	computations to be performed as compared to when original image is
	used as input image). When integrating this algorithm within a UAV
	drone equipped with a 720p camera, the feed recorded would be in the
	form of a video. This video feed is broken into frames (images) so
	it can be compatible to be fed into the algorithm for classification.
	A 60 fps, 720p camera, records 60 frames of $1280\times720$ pixel
	size every second. All these 60 frames are not relevant as they are
	a mere copy of each other at different angles, considering that UAV
	is under motion. Fig. \ref{fig:Fig5} presents the proposed SegNet
	architecture for early wildfire detection (needs to be shifted to
	the end of SegNet Architecture). These frames prove to be useful for
	training of the model but not for execution. Thus, selecting a frame
	every 20 frames in a second, helps lower computational costs and speed
	up processing. This also provides time for additional supplemental
	programs on UAV, for example, fire alert signaling process to send
	results to respective authorities for any follow up action.
	
	\subsection{SegNet Architecture}
	
	This algorithm is based on Convolution Neural Networks with Support
	Vector Machine used to draw a hyperplane among the data-points with
	ReLu classifier for convolutions \cite{Ref3,Ref12,Ref24,Ref26}. Successful
	use of linear support vector machines, for binary classification,
	was performed by many researchers \cite{Ref11,Ref12,Ref24,Ref29}.
	ReLu classifier has proven to be much faster at convergence compared
	to tanh and sigmoid \cite{Ref19}. The optimal hyperplane to separate
	two classes can be computed by the following set of equations \cite{Ref30}.
	\begin{equation}
		f(w,x)=w.x+b\label{eq:Sec32_eq1}
	\end{equation}
	\begin{equation}
		\text{min}\frac{1}{p}w^{\top}w+C\sum_{i=1}^{p}\text{max}(0,1-y_{i}(w^{t}x_{i}+b))^{2}\label{eq:Sec32_eq2}
	\end{equation}
	$w^{\top}w$ refers to the Manhattan norm (also known as $\mathcal{L}_{1}$
	norm), $C$ denotes penalty parameter, $y_{i}$ denotes the actual
	label, and $w^{t}x_{i}+b$ denotes the predictor function. The differentiable
	counterpart, $\mathcal{L}_{2}$-SVM provides more stable results \cite{Ref27}
	\begin{equation}
		\text{min}\frac{1}{p}||w||_{2}^{2}+C\sum_{i=1}^{p}\text{max}(0,1-y_{i}(w^{t}x_{i}+b))^{2}\label{eq:Sec32_eq3}
	\end{equation}
	The model comprises of five layers of convolutions and two hidden
	layers separated by a flatten layer with one output layer that has
	a single dense node. These, in an order, are outlined in Table \ref{tab:SegNet}.
	
	\begin{table}
		
		\caption{\label{tab:SegNet}SegNet architecture.}
		
		\begin{tabular}{c>{\raggedright}p{7.4cm}}
			\toprule 
			Layer & Description\tabularnewline
			\midrule
			\midrule 
			1 & 1st Convolution 2D layer: Filters = 32, Activation function = ReLu,
			kernel size = $3\times3$, Stride = 2, Input Shape = (240, 320, 3),
			padding = “SAME”\tabularnewline
			\midrule 
			2 & 1st Maxpool layer: Pool Size = 2, Stride = 2 \tabularnewline
			\midrule 
			3 & 2nd Convolution 2D layer: Filters = 64, Activation function = ReLu,
			kernel size = $3\times3$ , Stride = 2, padding = “SAME”\tabularnewline
			\midrule 
			4 & 2nd Maxpool layer: Pool Size = 2, Stride = 2 \tabularnewline
			\midrule 
			5 & 3rd Convolution 2D layer: Filters = 128, Activation function = ReLu,
			kernel size = $3\times3$, Stride = 2, padding = “SAME”\tabularnewline
			\midrule 
			6 & Flatten layer\tabularnewline
			\midrule 
			7 & 1st Dense layer: Units = 16, Activation function = ReLu \tabularnewline
			\midrule 
			8 & 2nd Dense layer: Units = 16, Activation function = ReLu \tabularnewline
			\midrule 
			9 & Output layer: Unit = 1, $\mathcal{L}_{2}$ regularization technique,
			Activation function = linear \tabularnewline
			\bottomrule
		\end{tabular}
	\end{table}

	\section{Workflow, Challenges, and Mitigation\label{sec:WorkFlow}}
	
	\subsection{Overfitting Issue}
	
	During the initial algorithm tests, a concerning phenomenon emerged:
	the model exhibited signs of overfitting, where it excessively tailored
	itself to the training data. To address this issue, data augmentation
	techniques were implemented, intending to introduce variety into the
	dataset and curb overfitting. While these methods did alleviate the
	issue to some degree, they did not completely eradicate the problem.
	Additionally, attempts to mitigate overfitting through early stopping
	proved ineffective as the root cause was not solely over-training.
	A breakthrough came with the application of $\mathcal{L}_{2}$ regularization.
	This technique proved to be remarkably effective in addressing the
	overfitting challenge. $\mathcal{L}_{2}$ regularization operates
	by penalizing overly complex models, in particular in terms of weight
	values, effectively balancing the weights associated with different
	nodes in the neural network. By doing so, it corrected the bias present
	in the initial weight distribution, ensuring that no specific node
	dominated the others. This balance in weights significantly improved
	the model’s generalization abilities, allowing it to perform better
	on unseen data. The implementation of $\mathcal{L}_{2}$ regularization
	is pivotal during the training process. It not only highlights the
	bias issue within the model but also provides a viable solution by
	equalizing the influence of various nodes. This corrective measure
	significantly enhances the model's ability to generalize, making it
	more reliable and effective in handling diverse datasets. Therefore,
	$\mathcal{L}_{2}$ regularization bolstering the overall robustness
	of the algorithm.
	
	\subsection{Workflow}
	
	\begin{figure*}
		\centering{}\includegraphics[scale=0.32]{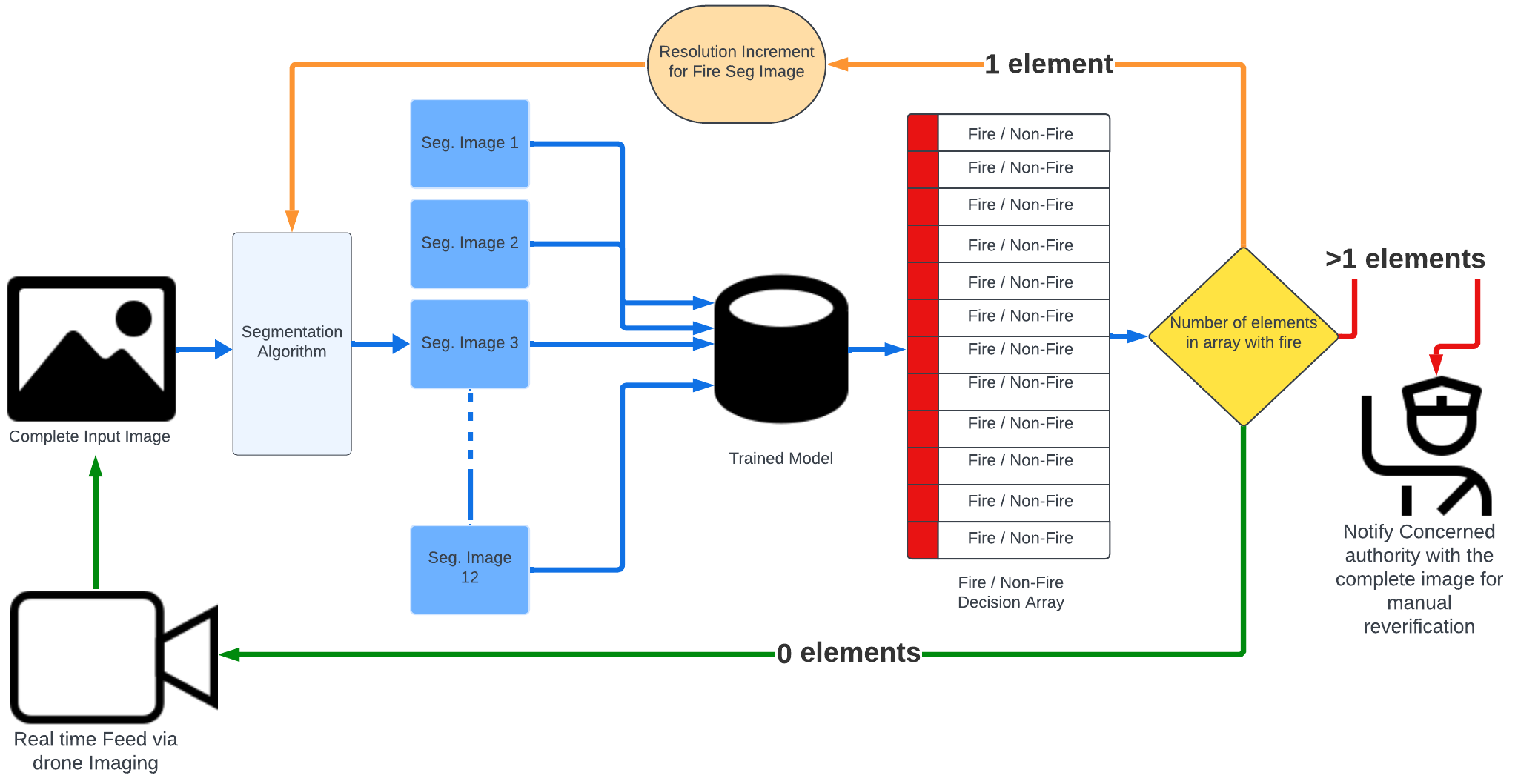}\caption{Workflow of a drone’s detection processing system with a model trained
			on the proposed SegNet architecture.}
		\label{fig:Fig6}
	\end{figure*}
	
	The development of this machine learning algorithm was meticulously
	designed with a specific workflow, as shown in Fig. \ref{fig:Fig6},
	to achieve accurate classification results. At its core, the algorithm
	processes live video feed captured by drones, a crucial component
	in wildfire surveillance. The process begins by temporarily storing
	the video feed in the drone's memory. This video stream is then divided
	into frames, with a specific subset of frames selected at fixed intervals,
	precisely one frame after every 19 frames in a 60 fps camera, translating
	to 3 frames per second. This intentional selection provides a buffer
	period, facilitated by the model's swift processing speeds, allowing
	for additional secondary tasks to be executed. These tasks include
	intricate processes such as segmentation algorithms, real-time notifications
	to authorities upon fire detection, control of the drone's flight
	systems, and integration with other sensors like GPS. One of the algorithm’s
	remarkable features is its ability to optimize resource utilization.
	By incorporating these secondary tasks within the main onboard processing
	unit, it eliminates the necessity for parallel processing units, reducing
	both the weight and production costs of surveillance drones significantly.
	This streamlined approach ensures the algorithm operates seamlessly
	within the drone’s existing framework, enhancing efficiency without
	compromising the drone's overall functionality.
	
	From an engineering perspective, this design choice not only minimizes
	the drone's weight but also curtails production expenses, a critical
	factor given the need for deploying a substantial number of drones
	to cover vast forested areas effectively. The algorithm achieves this
	efficiency by employing a simple yet effective architecture. It utilizes
	a minimal number of convolution and dense layers, strategically balancing
	computational costs with high detection accuracy and rapid processing
	speeds. Upon receiving the segmented frames, depicted in Fig. \ref{fig:Fig6},
	the algorithm proceeds to the segmentation stage. Each frame is dissected
	into 12 distinct segments, each maintaining a pixel size of $320\times240$
	without any loss of critical information. These individual segments
	are then fed into the machine learning model, meticulously trained
	for fire detection. The model's output, whether it signifies the presence
	of fire or absence thereof, is stored in what is termed a \textquotedbl decision
	array\textquotedbl{} within the system. The decision array serves
	as a pivotal component in the algorithm's decision-making process.
	If the array contains zero elements classified as fire, the algorithm
	accepts a new image from the live video feed storage, initiating a
	continuous cycle of assessment. When the array contains one element
	categorized as fire, the corresponding segmented image is reprocessed.
	Before reprocessing, this segment is rescaled to match the dimensions
	of a complete image. This step is fundamental in validating the initial
	classification, preventing potential false positives, and ensuring
	precision in the detection process. However, when the decision array
	reveals two or more elements marked as fire, the drone triggers a
	significant action. The complete image under examination is promptly
	forwarded to the relevant authorities for comprehensive verification
	and necessary further action. This approach ensures that potential
	wildfire incidents are promptly and accurately reported, allowing
	authorities to take swift and informed measures in response to the
	detected threat.
	
	\section{Results\label{sec:Results}}
	
	\subsection{SegNet Performance and Test Accuracy }
	
	The personal computing hardware used for building this model are:
	GPU -- GTX 1660 TI MaxQ design (192-bit memory interface, 1.14 GHz
	clock speed), CPU -- AMD Ryzen 9 4900HS (3 GHz clock speed, 7 nm
	process size), Software -- Python 3.11.4, TensorFlow, Keras, numpy,
	PILLOW. An approach marked by caution was adopted, ensuring that enhancements
	were made conservatively to preserve the algorithm’s processing speed
	when it came to hyperparameter tuning and tuning the number of convolution
	and dense layers. Please follow this \href{https://youtu.be/xwMzFpZkC8M}{URL}
	for video of the experiment.
	
	\begin{table}[h]
		\caption{\label{tab:ComprehensiveConv}Comprehensive performance analysis of
			the proposed SegNet: (a) Processing times of Segmented approach model
			with varying convolution and dense layers, and (b) Test accuracy of
			Segmented approach model with varying convolution and dense layers.}
		
		\begin{tabular}{cccccc}
			\hline 
			\noalign{\vskip\doublerulesep}
			\multirow{2}{*}{} & \multicolumn{5}{c}{SegNet number of convolution layers, time, and accuracy}\tabularnewline[\doublerulesep]
			\cline{2-6} \cline{3-6} \cline{4-6} \cline{5-6} \cline{6-6} 
			\noalign{\vskip\doublerulesep}
			& 5 Conv & 6 Conv & 7 Conv & 9 Conv & 11 Conv\tabularnewline[\doublerulesep]
			\hline 
			\hline 
			\noalign{\vskip\doublerulesep}
			1 Dense & Low Acc & $389.32$ ms & $566$ ms & $812$ ms & Low Acc\tabularnewline[\doublerulesep]
			\cline{2-6} \cline{3-6} \cline{4-6} \cline{5-6} \cline{6-6} 
			\noalign{\vskip\doublerulesep}
			Layer & $54.71\%$ & $97.5\%$ & $96.82\%$ & $96.14\%$ & $48.79\%$\tabularnewline[\doublerulesep]
			\hline 
			\noalign{\vskip\doublerulesep}
			2 Dense & $240.37$ ms & $1.34$ sec & $2.91$ sec & $3.32$ sec & Low Acc\tabularnewline[\doublerulesep]
			\cline{2-6} \cline{3-6} \cline{4-6} \cline{5-6} \cline{6-6} 
			\noalign{\vskip\doublerulesep}
			Layer & $98.18\%$ & $98.18\%$ & $97.05\%$ & $95.23\%$ & $51.24\%$\tabularnewline[\doublerulesep]
			\hline 
			\noalign{\vskip\doublerulesep}
			3 Dense & $2.95$ sec & $4.06$ sec & $6.47$ sec & Low Acc & Low Acc\tabularnewline[\doublerulesep]
			\cline{2-6} \cline{3-6} \cline{4-6} \cline{5-6} \cline{6-6} 
			\noalign{\vskip\doublerulesep}
			Layer & $97.95\%$ & $97.7\%$ & $94.78\%$ & $72.56\%$ & $39.31\%$\tabularnewline[\doublerulesep]
			\hline 
		\end{tabular}
	\end{table}
	
	In the developmental stages of the model, depicted in Table \ref{tab:ComprehensiveConv},
	various configurations of convolution layers (ranging from 5 to 11
	layers) and dense layers (ranging from 1 to 3 layers) were systematically
	tested. Note that Conv is a short abbreviation denotes convolution.
	Prioritizing computational efficiency, a balance was sought between
	model speed and accuracy. Initial experiments with 5 convolution layers
	and 1 dense layer yielded a suboptimal accuracy of $54.71\%$, while
	an extensive configuration of 11 convolution layers and 3 dense layers
	resulted in a lower accuracy of $39.31\%$. The optimal compromise
	between accuracy and processing speed emerged with a configuration
	of 5 convolution layers and 2 dense layers, achieving an impressive
	$98.18\%$ accuracy and a processing time of $240.375$ ms. This careful
	exploration of model architecture during development ensures a harmonious
	blend of computational efficiency and accuracy in the proposed research.
	
	\begin{table}[h]
		\caption{\label{tab:Analysis}SegNet imaging approaches.}
		
		\begin{tabular}{c>{\centering}p{1.2cm}>{\centering}p{1.4cm}>{\centering}p{1.4cm}>{\centering}p{1.5cm}}
			\cline{2-5} \cline{3-5} \cline{4-5} \cline{5-5} 
			\noalign{\vskip\doublerulesep}
			\multirow{2}{*}{} & \multicolumn{4}{c}{Dataset Categorization}\tabularnewline[\doublerulesep]
			\cline{2-5} \cline{3-5} \cline{4-5} \cline{5-5} 
			\noalign{\vskip\doublerulesep}
			& Original Complete Images & Augmented Complete Images & Segmented Images & Total\tabularnewline[\doublerulesep]
			\hline 
			\hline 
			\noalign{\vskip\doublerulesep}
			Fire & $5,195$ & $547$ & $2,266$ & $5,968$\tabularnewline[\doublerulesep]
			\hline 
			\noalign{\vskip\doublerulesep}
			Non-Fire & $2,117$ & $239$ & $1,918$ & $4,274$\tabularnewline[\doublerulesep]
			\hline 
			\noalign{\vskip\doublerulesep}
			Total & $7,312$ & $786$ & $2,144$ & $10,242$\tabularnewline[\doublerulesep]
			\hline 
		\end{tabular}
	\end{table}
	
	\begin{table}[h]
		\caption{\label{tab:Enhance}SegNet different training consideration and enhancements.}
		
		\begin{tabular}{c>{\centering}p{1.1cm}>{\centering}p{1.2cm}>{\centering}p{1.7cm}>{\centering}p{1.5cm}}
			\cline{2-5} \cline{3-5} \cline{4-5} \cline{5-5} 
			\noalign{\vskip\doublerulesep}
			& \multicolumn{4}{>{\centering}p{6.7cm}}{SegNet model accuracies with different training considerations and
				enhancements}\tabularnewline[\doublerulesep]
			\cline{2-5} \cline{3-5} \cline{4-5} \cline{5-5} 
			\noalign{\vskip\doublerulesep}
			& Simple Training & Early Stopping & Data Augmentation & $\mathcal{L}_{2}$-regularization\tabularnewline[\doublerulesep]
			\hline 
			\hline 
			\noalign{\vskip\doublerulesep}
			Training & $89.2\%$ & $89.3\%$ & $97.5\%$ & $99.6\%$\tabularnewline[\doublerulesep]
			\hline 
			\noalign{\vskip\doublerulesep}
			Validation & $59.7\%$ & $59.3\%$ & $76.1\%$ & $99.3\%$\tabularnewline[\doublerulesep]
			\hline 
			\noalign{\vskip\doublerulesep}
			Testing & $56.7\%$ & $56.9\%$ & $71.9\%$ & $98.2\%$\tabularnewline[\doublerulesep]
			\hline 
		\end{tabular}
	\end{table}
	
	Throughout the developmental phases, as shown in Table \ref{tab:Analysis}
	and Table \ref{tab:Enhance}, the algorithm's performance was carefully
	scrutinized, with model test accuracy being systematically recorded
	at each stage of improvement. Table \ref{tab:Analysis} lists number
	of images in the training dataset including original images, augmented
	complete images, and segmented images for the case of fire and non-fire.
	Table \ref{tab:Enhance} illustrates accuracy improvements with different
	techniques to deal with overfitting. In its initial iteration, the
	model exhibited a $56.7\%$ accuracy, unmitigated by overfitting concerns.
	Notably, the training accuracy stood at $89.2\%$, while the validation
	accuracy was slightly higher at $59.7\%$ after 15 epochs. To counter
	overfitting, the first strategy employed was early stopping. However,
	this measure yielded results that were remarkably similar to the previous
	state, indicating that it did not substantially mitigate the overfitting
	issue. Subsequently, the integration of augmented images into the
	dataset emerged as the next step. This strategic addition led to a
	significant boost in training accuracy, elevating it to $97.5\%$,
	accompanied by a substantial rise in validation accuracy to $76.1\%$.
	Consequently, the model's test accuracy, evaluated using a dataset
	comprising $441$ images, experienced a noteworthy increase from $56.7\%$
	to $71.8\%$. Recognizing the need for a better solution, $\mathcal{L}_{2}$
	norm regularization was implemented. This technique played a pivotal
	role in refining the model by adjusting weights and mitigating biases
	towards specific features. The impact was profound, evident in the
	substantial increase in training accuracy to $99.6\%$, with a corresponding
	validation accuracy of $99.3\%$. This adjustment resulted in a remarkable
	enhancement of the model's test accuracy, reaching $98.2\%$. The
	cautious integration of techniques, from early stopping to augmented
	data, and finally, $\mathcal{L}_{2}$ norm regularization, resulted
	in a finely-tuned model. By addressing overfitting and refining its
	generalization capabilities, the algorithm achieved an accuracy rate
	that underlines its effectiveness and reliability, making it a potent
	tool in the realm of wildfire detection.
	
	\subsection{Comparison Cutting Edge Techniques}
	
	\begin{figure*}
		\centering{}\includegraphics[scale=0.35]{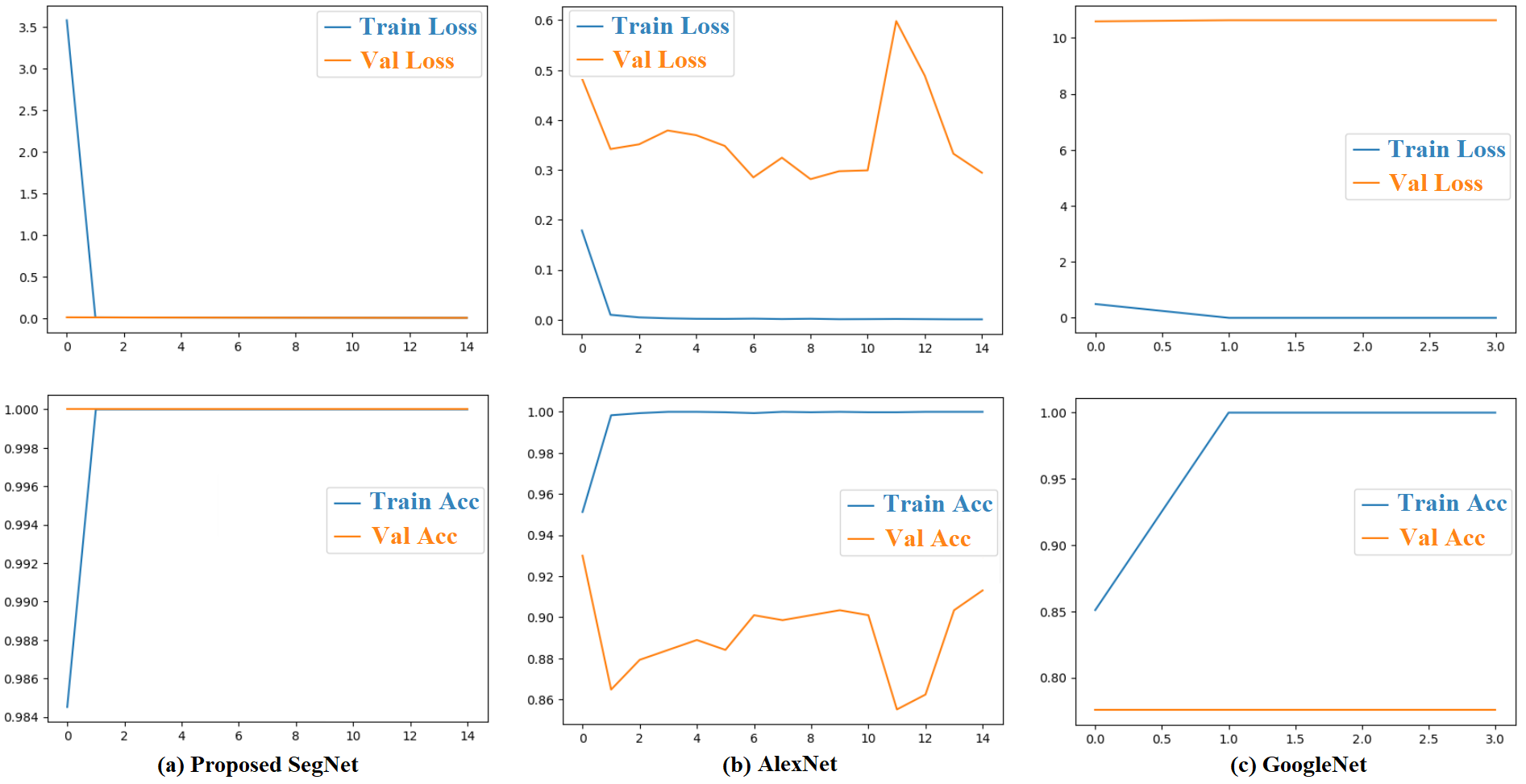}
		
		\caption{Comparison of Training/Validation accuracy and Losses to state-of-the-art
			literature: (a) Proposed SegNet, (b) AlexNet (literature), and (c)
			GoogleNet (literature).}
		\label{fig:Fig7}
	\end{figure*}
	
	To better understand the effectiveness of this proposed segmentation
	approach, the results are compared with state-of-the-art literature
	algorithms. These algorithms are GoogleNet \cite{Ref38} and AlexNet
	\cite{Ref10}. Both the algorithms were trained with the same dataset
	excluding the segmented images to help with proper comparison of results.
	Fig. \ref{fig:Fig7}.(a) shows the proposed SegNet performance, with
	train accuracy of $99.6\%$, validation accuracy of $99.3\%$, and
	test accuracy of $98.2\%$ with $0.0003$ train loss and validation
	loss of $0.0004$. Fig. \ref{fig:Fig7}.(b) presents AlexNet achieved
	model train accuracy of $100\%$, test accuracy of $92.2\%$, and
	validation accuracy of $91.3\%$ with 0.0005 training loss and $0.2943$
	validation loss. Fig. \ref{fig:Fig7}.(c) depicts GoogleNet achieved
	model train accuracy of $99.9\%$, test accuracy of $76.8\%$ and
	validation accuracy of $77.5\%$. The recorded losses for training
	were $0.24$ and an increased loss $10.629$. 
	
\begin{figure}
		\centering{}\includegraphics[scale=0.18]{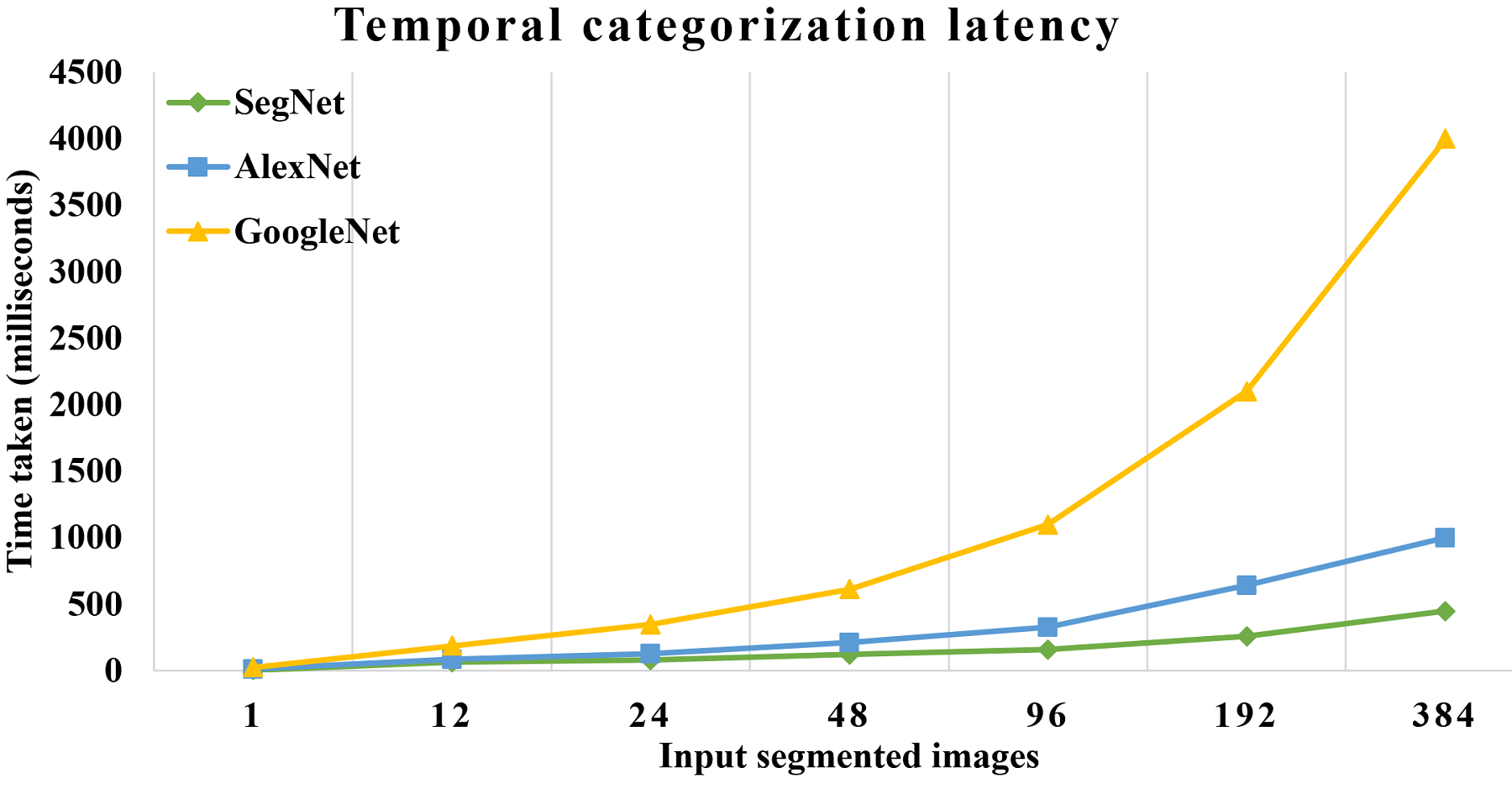}\caption{Temporal Categorization Latency chart: This chart represents the time
			required to perform classification task with different batch sizes
			of input images.}
		\label{fig:Fig9}
	\end{figure}

	One of the main points of focus for this research was to reduce processing
	times along with the use of minimal computational power. Please visit
	\nameref{subsec:Appendix-A} for comprehensive discussion about performance
	and time complexities. The processing speeds for proposed segmentation
	approach were observed and recorded at the stage of hyperparameter
	tuning and selection of number of convolution and dense layers. For
	the segmented approach single batch of 32 segmented images took an
	average of $641$ milliseconds.
	
	\[
	\frac{641\text{ ms}}{32\text{ Seg Images}}=20.0312\frac{\text{ms}}{\text{Seg Images}}
	\]
	A complete image consists of $12$ Segmented images:
	\[
	20.0312\frac{\text{ms}}{\text{Seg Images}}\times12=240.375\frac{\text{ms}}{\text{complete image}}
	\]
	Thus, in the proposed segmented approach, a complete image of $1280\times720$
	pixel resolution was passed through the neural network within 240.37
	milliseconds. The processing speeds of SegNet approach, GoogleNet
	and AlexNet were also compared for better understanding. As all the
	convolution layers are interconnected in GoogleNet in the form of
	inception block, which results in high number of computations, GoogleNet
	architecture was found to be the slowest with $1.487$ seconds. AlexNet
	performed better than GoogleNet as it was observed to be faster at
	$661.54$ milliseconds. The proposed SegNet was observed to be faster
	than both GoogleNet and AlexNet with $240.375$ milliseconds per complete
	image. The times mentioned are the average time observed for one complete
	image over 10 trials. As shown on Fig. \ref{fig:Fig9}, SegNet algorithm
	performs the best among the three models for each value of batch size.
	GoogleNet’s curve represents an incline in latency with increase in
	batch size. This is due to the interconnections of various different
	convolution blocks in the GoogleNet model architecture. Whereas, both
	AlexNet and SegNet perform with a curve close to linearity.
	
	\subsection{SegNet Memory Requirements and Time Complexities}
	
\begin{figure}[h]
		\centering{}\subfloat[]{\includegraphics[scale=0.2]{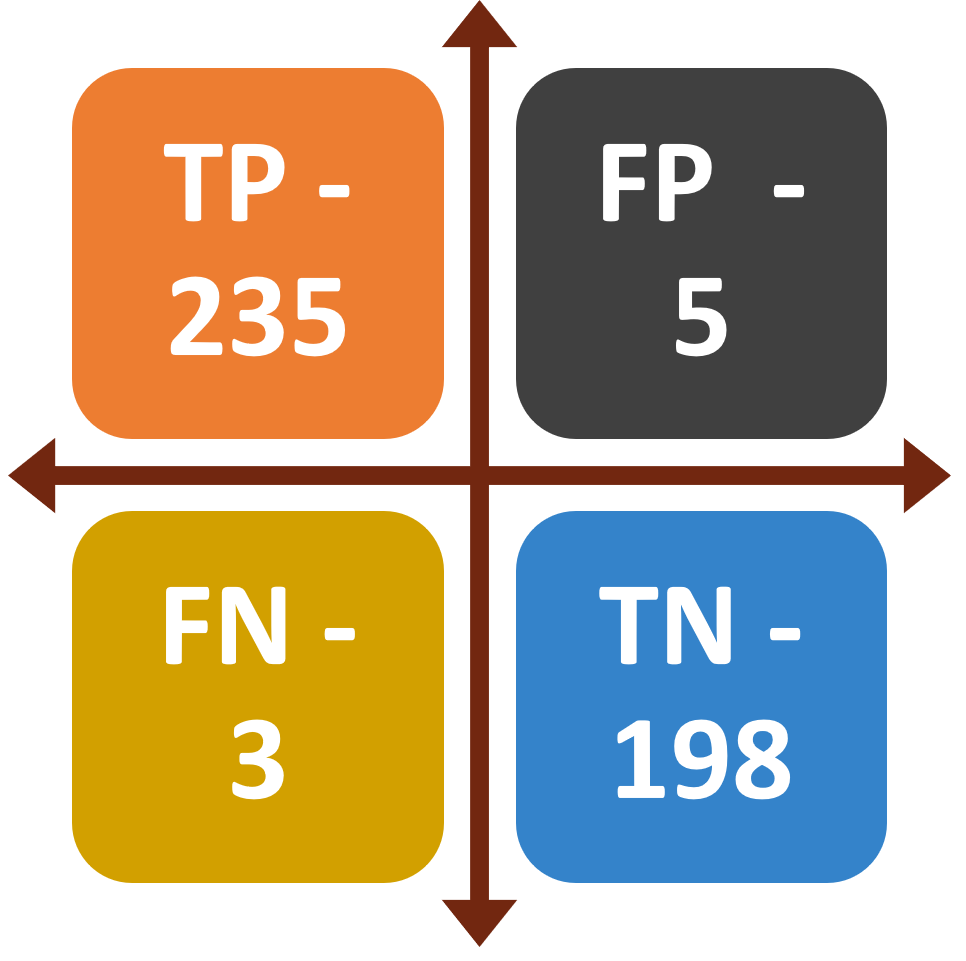}} 
		
		\subfloat[]{\includegraphics[scale=0.17]{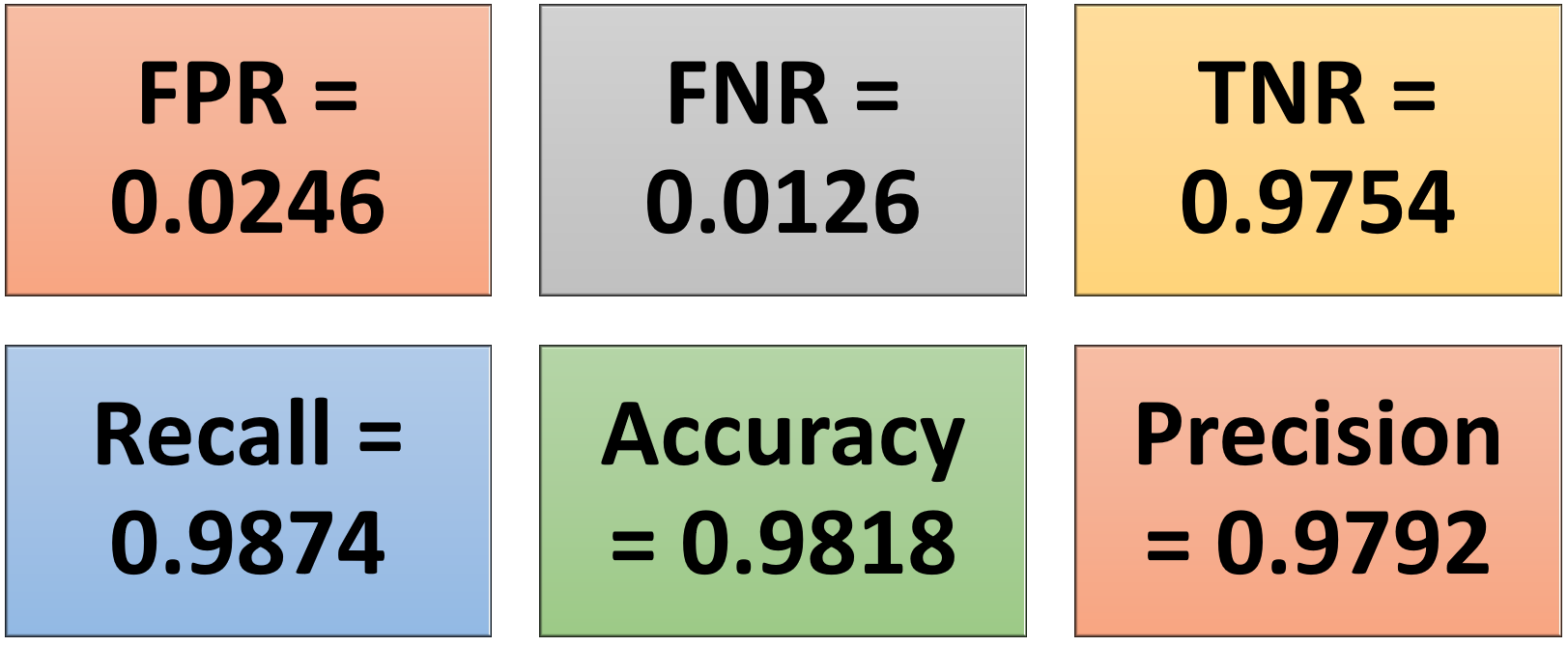}}
		
		\caption{SegNet performance metric: (a) True Positive (TP), False Positive
			(FP), False Negative (FN), True Negative (TN); and (b) False Positive
			Rate (FPR), False Negative Rate (FNR), True Negative Rate (TNR), recall,
			accuracy and precision of the model.}
		\label{fig:Fig8}
	\end{figure}

	Memory requirements and time complexities calculations are detailed
	in \nameref{subsec:Appendix-A}. The proposed SegNet architecture’s
	performance metrics are presented in Fig. \ref{fig:Fig8}.(a) and
	Fig. \ref{fig:Fig8}.(b) in terms of accuracy and precision rates.
	Fig. \ref{fig:Fig8}.(a) shows the number of images that were classified
	as one of the mentioned categories by SegNet on test data. Out of
	the selected 441 complete images for testing, 235 were categorized
	correctly as True Positives (TPs) and 198 were classified as True
	Negatives (TNs). This testifies to SegNet model’s effectiveness. As
	depicted, 5 False Positives (FPs) and 3 True Negatives signify the
	imperfections found in the model. With Accuracy of $98.18\%$ that
	exceeds results obtained by AlexNet and GoogleNet of $92.2\%$ and
	$76.8\%$ respectively. Fig \ref{fig:Fig8}.(b) shows the different
	performance metrics of SegNet. With a high recall, precision and accuracy,
	SegNet performs on par with most modern machine learning approaches,
	along with the model being computationally faster.
	
	Table \ref{tab:Memory} shows the memory consumption of the SegNet
	model. The SegNet model consumes 44.17 megabytes or 44,177,408 bytes
	of memory. Thus, any computational device on board a drone, with 50
	megabytes of additional memory can perform operations of SegNet model.
	Table 6 also represents the total number of operations carried out
	for every layer. The total number of operations for SegNet model are
	19.275 million. Thus, a drone with a processor with 19.275 MHz speed
	is required. For example, a common CPU on board drones manufactured
	by Qualcomm, the snapdragon 821 has the ability to perform at 2.15
	GHz which could prove ideal for both the drone flight and wildfire
	detection algorithm’s processes. The data in Table \ref{tab:Memory}
	were collected using the expressions in \nameref{subsec:Appendix-A}.
	
	\begin{table}[h]
		\caption{\label{tab:Memory}Metrics performance and the related mathematical
			formulas.}
		
		\begin{tabular}{>{\raggedright}p{1.3cm}>{\raggedright}p{1.3cm}>{\raggedright}p{1.4cm}>{\raggedright}p{1.4cm}>{\raggedright}p{1.3cm}}
			\hline 
			\noalign{\vskip\doublerulesep}
			Layers & Parameters & Model space requirement (bytes) & Operational space requirement (bytes) & Total operations performed\tabularnewline[\doublerulesep]
			\hline 
			\hline 
			\noalign{\vskip\doublerulesep}
			Conv2D 1 & 896 & 7,168 & 2,889,856 & 8,242,560\tabularnewline[\doublerulesep]
			\hline 
			\noalign{\vskip\doublerulesep}
			MaxPool 1 & - & - & -  & -\tabularnewline[\doublerulesep]
			\hline 
			\noalign{\vskip\doublerulesep}
			Conv2D 1 & 18,496 & 147,968 & 1,322,752 & 4,095,360\tabularnewline[\doublerulesep]
			\hline 
			\noalign{\vskip\doublerulesep}
			MaxPool 1 & - & - & -  & -\tabularnewline[\doublerulesep]
			\hline 
			\noalign{\vskip\doublerulesep}
			Conv2D 1 & 73,856 & 590,848 & 636,544 & 2,021,760\tabularnewline[\doublerulesep]
			\hline 
			\noalign{\vskip\doublerulesep}
			Flatten & - & - & -  & -\tabularnewline[\doublerulesep]
			\hline 
			\noalign{\vskip\doublerulesep}
			Dense 1 & 2,457,616 & 18,530,432 & 39,322,368  & 4,915,200\tabularnewline[\doublerulesep]
			\hline 
			\noalign{\vskip\doublerulesep}
			Dense 2 & 272 & 2,176 & 4,864 & 512\tabularnewline[\doublerulesep]
			\hline 
			\noalign{\vskip\doublerulesep}
			Output & 17 & 136 & 1,024 & 32\tabularnewline[\doublerulesep]
			\hline 
		\end{tabular}
	\end{table}

	\section{Conclusion \label{sec:Conclusion}}
	
	In conclusion, this research addresses the critical challenges in
	wildfire detection, focusing on enhancing time resolution and optimizing
	processing speeds while maintaining high accuracy levels. Leveraging
	the power of Convolutional Neural Networks (CNNs), the study introduced
	a new approach: Segmented Neural Network (SegNet). This innovative
	method involved breaking down high-resolution images into smaller,
	manageable segments, allowing for rapid processing without compromising
	accuracy. The study meticulously navigated the complexities of feature
	reduction, acknowledging the inherent difficulties in balancing extensive
	feature maps with limited datasets. Through a systematic process,
	various techniques were applied to mitigate overfitting, enhance generalization,
	and achieve outstanding accuracy rates. Early stopping, data augmentation,
	and $\mathcal{L}_{2}$ norm regularization were sequentially implemented,
	each step refining the algorithm's capabilities. The results demonstrated
	a significant leap in accuracy, with the final model achieving an
	impressive $98.2\%$ accuracy on the test dataset. A pivotal aspect
	of this research is the comparison with established algorithms, GoogleNet
	and AlexNet, which provided valuable insights. The segmentation approach
	outperformed these models, showcasing not only superior accuracy but
	also remarkable processing speeds. The segmentation technique processed
	a complete image of $1280\times720$ pixels in just $240.37$ milliseconds,
	a testament to its efficiency in real-time applications. Beyond accuracy
	and speed, this research emphasized the importance of preserving essential
	details in images, especially in early wildfire detection scenarios.
	By ensuring the amorphous nature of fire, features were retained through
	segmentation, the algorithm excelled in detecting subtle signs of
	wildfire, crucial for timely intervention.
	
	Furthermore, this research highlighted the algorithm's integration
	within UAV drone systems. By optimizing resource utilization and minimizing
	computational costs, the algorithm can seamlessly integrate secondary
	tasks, such as real-time notifications, segmentation algorithms, and
	control systems, without the need for additional processing units.
	This streamlined approach significantly reduces production costs and
	drone weight, ensuring practicality and efficiency in real-world deployments.
	In summary, this research not only introduces a new segmentation approach
	for wildfire detection but also presents a comprehensive framework
	that balances accuracy, speed, and efficiency. By addressing critical
	challenges in real-time processing and detection accuracy, this study
	contributes significantly to the advancement of wildfire surveillance
	systems, promising enhanced safety and rapid response capabilities
	in the face of wildfire threats. 
	
	\section*{Acknowledgment}

	\subsection*{Appendix A\label{subsec:Appendix-A}}
	\begin{center}
		\textbf{Calculation of Performance and Time Complexities}
		\par\end{center}
	
	\noindent Let us define True Positives (TP) as the number of fire
	images correctly classified as fire images by the model, True Negatives
	(TN) as the number of non-fire images correctly classified as non-fire
	images by the model, False Positives (FP) as number of non-fire images
	classified as fire images by the model, and False Negatives (FN) as
	number of fire images classified as non-fire images by the model.
	The mathematical expressions for the performance metrics presented
	in Fig. \ref{fig:Fig8} follows Table \ref{tab:Metrics}.
	
	\begin{table}[h]
		\caption{\label{tab:Metrics}Metrics performance and the related mathematical
			formulas.}
		
		\begin{tabular}{>{\raggedright}p{1.7cm}>{\raggedright}p{2.1cm}>{\raggedright}p{3.8cm}}
			\hline 
			\noalign{\vskip\doublerulesep}
			Metrics performance & Mathematical Formula & Interpretation\tabularnewline[\doublerulesep]
			\hline 
			\hline 
			\noalign{\vskip\doublerulesep}
			False Positive Rate (FPR) & $\frac{FP}{FP+TN}$ & Measure of how incorrect the model is classifying fire image\tabularnewline[\doublerulesep]
			\hline 
			\noalign{\vskip\doublerulesep}
			False Negative Rate (FNR) & $\frac{FN}{TP+FN}$ & Measure of how incorrect the model is classifying non-fire image\tabularnewline[\doublerulesep]
			\hline 
			\noalign{\vskip\doublerulesep}
			True Negative Rate (TNR) & $\frac{TN}{FP+TN}$ & Measure of non-fire image correctness of the model\tabularnewline[\doublerulesep]
			\hline 
			\noalign{\vskip\doublerulesep}
			Recall & $\frac{TP}{TP+FN}$ & Measure of fire images correctness of the model\tabularnewline[\doublerulesep]
			\hline 
			\noalign{\vskip\doublerulesep}
			Accuracy & $\frac{TP+TN}{TP+FN+TN+FP}$ & Overall effectiveness of the model to classify correctly\tabularnewline[\doublerulesep]
			\hline 
			\noalign{\vskip\doublerulesep}
			Precision & $\frac{TP}{TP+FP}$ & The accuracy of positive predictions\tabularnewline[\doublerulesep]
			\hline 
		\end{tabular}
	\end{table}
	
	\paragraph*{Time complexity}The time complexity analysis for the
	proposed SegNet algorithm, AlexNet and GoogleNet have been performed
	with Big-O asymptotic notation. Let us start by defining the following
	variables: $|w|$ refers to number of weight elements, $c\times w\times h$
	describes Kernel size and channels, $k$ denotes number of Kernels,
	$m_{o}$ denotes number of neurons in output dense layer, $p$ is
	the memory size for float64 datatype element (8 bytes), $s$ denotes
	strides, $M\times N$ denotes the input image dimensions, $n$ denotes
	neurons in input layer in dense layers. Big-O notation $O(g(n))$
	is defined as follows: 
	\begin{align}
		f(n)= & O(g(n))\text{ iff }\exists\text{ positive constant }c\text{ and }n_{0}\nonumber \\
		& \text{\ensuremath{\hspace{1em}}such that }f(n)\leq c\times g(n)\forall n\geq n_{0}\label{eq:append_eq1}
	\end{align}
	$n$ denotes number of inputs, $n_{0}$ denotes a positive integer,
	$w\times h$ denotes Kernel size, $c$ denotes number of channels
	\cite{Ref51}. For change in image size, the total number of operations
	per convolution layer ($TOpL$) in SegNet model is given by
	\begin{equation}
		TOpL=2\frac{(c\times w\times h)(M-w+s)(N-h+s)}{s^{2}}\label{eq:append_eq2}
	\end{equation}
	Since the only affected factors with change in image size are dimensions
	$M$ and $N$, the above equation can be simplified as follows:
	\begin{align}
		f(n) & =c\times g(n)\nonumber \\
		& =G\times(M-a)\times(N-b)\nonumber \\
		G & =2\frac{c\times w\times h}{s^{2}}\label{eq:append_eq3}
	\end{align}
	where $a=w-s$ and $b=h-s$. Thus, the time complexity with respect
	to change in image size is linear in two dimensions $M$ and $N$.
	The Big-O notation for this algorithm is $O(M,N)$.
	
	\paragraph*{Memory requirements and operational complexity}The model
	space complexity is defined as the total amount of memory utilized
	by the model in a given computational environment. Operational Space
	complexity presents the amount of memory being utilized by each layer’s
	operation. This metric of performance is used to determine the local
	efficiency of the layers in the model in this deep learning algorithm.
	Operational compute complexity describes the total number of operations
	being carried out in a convolutional neural network. These metrics
	can be determined by the following mathematical expressions. For the
	SegNet model \cite{Ref53}
	\begin{equation}
		|w|=k(c\times w\times h+1)\times p\,\text{bytes}\label{eq:append_eq4}
	\end{equation}
	The Input Image Memory ($IIM$) is defined by
	\begin{equation}
		IIM=c\times M\times N\times p\,\text{bytes}\label{eq:append_eq5}
	\end{equation}
	The Generated Output Memory ($GOM$) is equal to Gradient of Activation
	(GoA) and expressed as
	\begin{equation}
		GOM=\frac{k(M-w+s)(N-h+s)}{s^{2}}\times p\,\text{bytes}\label{eq:append_eq6}
	\end{equation}
	In view of Equation \eqref{eq:append_eq4}, \eqref{eq:append_eq5},
	and \eqref{eq:append_eq6}, the Total Operational Space Complexity
	($TOSC$) per layer is defined as
	\begin{equation}
		TOSC=IIM+|w|+2GOM\label{eq:append_eq7}
	\end{equation}
	During dense layer operations, the space required for weight matrix
	and gradient of weights is equal to $nm_{o}+m_{o}$. The space occupied
	by input and output neurons and backpropagation is $2n+2m_{o}$. The
	total space occupied by dense layer ($TSpDL$) is defined by
	\begin{equation}
		TSpDL=(|\bigtriangledown w|+|w|+2n+2k)\times p\,\text{bytes}\label{eq:append_eq8}
	\end{equation}
	Number of multiplication or addition operations per layer ($NOpL$)
	is expressed as \cite{Ref49}
	\begin{equation}
		NOpL=\frac{(c\times w\times h)(M-w+s)(N-h+s)k}{s^{2}}\label{eq:append_eq9}
	\end{equation}
	Total operations per layer ($TOpL$) is defined as in \eqref{eq:append_eq5}.
	This implies the total number of operations $TOpL$ do not depend
	on the number of inputs $n$. Hence, the multiplication or addition
	operations per output neuron is equal to $n$, and the total operations
	for all output neurons in a layer is equal to $2nm_{0}$.

	\balance
	\bibliographystyle{IEEEtran}
	\bibliography{bib_Wild}
	
\end{document}